\documentclass[11pt,letterpaper]{article}
\usepackage{emnlp2016}
\usepackage{times}
\usepackage{latexsym}
\usepackage{url}
\usepackage{graphicx}
\usepackage{float}
\usepackage{amsmath}
\emnlpfinalcopy

\def\emnlppaperid{***}

\title{Neural Machine Translation with External Phrase Memory}

\author{Yaohua Tang \\ The University of Hong Kong \\  tangyh@hku.hk\\
         \And
         Fandong Meng \\Institute of Computing Technology\\Chinese Academy of Sciences\\ mengfandong@ict.ac.cn\\
         \AND
                  Zhengdong Lu \and Hang Li \\ Noah's Ark Lab, Huawei Technologies\\
                  Lu.Zhengdong@huawei.com\\
                  HangLi.HL@huawei.com\\
                  \And Philip L.H. Yu\\ plhyu@hku.hk  \\The University of Hong Kong\\                  
         }

\date{}

\begin{document}

\maketitle

\begin{abstract}
In this paper, we propose phraseNet, a neural machine translator with a phrase memory which stores phrase pairs in symbolic form, mined from corpus or specified by human experts. For any given source sentence, phraseNet scans the phrase memory to determine the candidate phrase pairs and integrates tagging information in the representation of source sentence accordingly. The decoder utilizes a mixture of word-generating component and phrase-generating component, with a specifically designed strategy to generate a sequence of multiple words all at once. The phraseNet not only approaches one step towards incorporating external knowledge into neural machine translation, but also makes an effort to extend the word-by-word generation mechanism of recurrent neural network. Our empirical study on Chinese-to-English translation shows that, with carefully-chosen phrase table in memory,  phraseNet yields 3.45 BLEU improvement over the generic neural machine translator. 
\end{abstract}

\section{Introduction}
Neural machine translation (NMT), although only proposed recently,  has shown great potential, and arguably surpassed statistical machine translation on tasks like English-German translation~\cite{sennrich2015improving:120}.  In addition to its superior ability in modeling the semantics of source sentence and language fluency of the target sentence, NMT as a framework also has remarkable flexibility in accommodating  other form of knowledge. 

In this paper, we explore the possibility of equipping regular neural machine translator with an external memory storing rules that specify phrase-level correspondence between the source and target languages. Those rules are in symbolic form, which can be either extracted from a parallel corpus or given by experts.  We tailor the encoder, decoder and the attention model of the neural translator to help locate phrases in the source and generate their translations in the target. The proposed model is called \text{phraseNet}. phraseNet is not only one step towards incorporating external knowledge in to neural machine translation, but also an effort to extend the word-by-word generation mechanism of recurrent neural network. 

\subsection{Model Overview}
\begin{center}
\begin{figure}[h!]
\centering
\includegraphics[height=4.9cm]{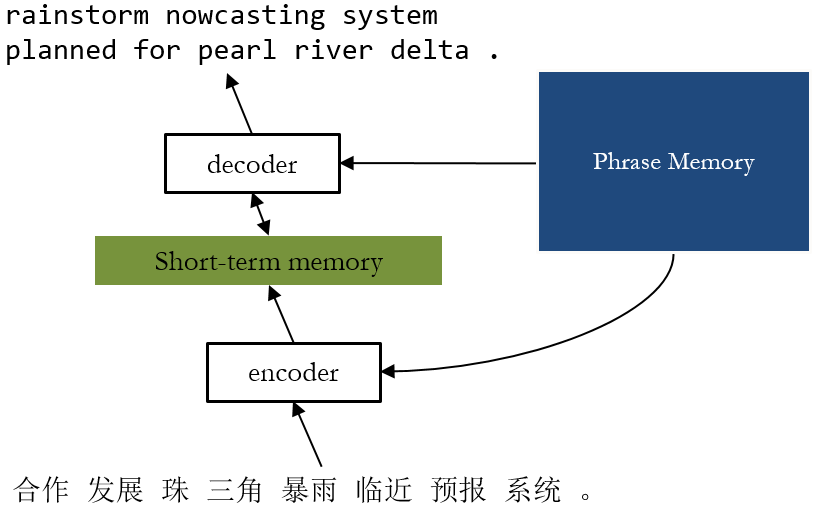}
\caption{The overall diagram of phraseNet.}
\label{fig-diagram}
\end{figure}
\end{center}

The overall diagram of phraseNet is displayed in Figure~\ref{fig-diagram}.
Basically, for a given source sentence, phraseNet first encodes the sentence to a representation with an RNN encoder, scans the phrase memory to select candidate phrase pairs and then tags the source representation accordingly (Section \ref{sec:encoder}). phraseNet then generates both words and phrases with an RNN decoder. It dynamically determines  at each time step with its probabilistic model consisting of a mixture of word-generation mode and phrase-generation mode (Section \ref{sec:models}). To maintain the state consistency of RNN when running in different modes, the decoder of phraseNet will go through ``idle run'' (Section \ref{sec:idlerun}) after generating a multiple-word phrase. \vspace*{-10pt}
\vspace*{-9pt}
Our contribution is of three-folds: 
\vspace{-5pt}
\begin{enumerate} 
\item we propose an end-to-end learning algorithm for neural machine translation with an external phrase memory, which to our knowledge it is the first effort in this direction; \vspace{-5pt}
\item we propose a way to handle the generation of multiple words with an RNN decoder;
\item our empirical studies on Chinese-English translation tasks show the efficacy of our models: phraseNet achieves on average 3.45 BLEU improvement over its generic counterpart.
\end{enumerate}

\paragraph{RoadMap} The remainder of this paper is organized in the following way. In Section 2, we will give a brief introduction to attention-based neural machine translation as the background. In Section 3, we will introduce phraseNet, including its two variants. In Section 4, we will report our experiments on applying phraseNet to  Chinese-English translation tasks. Then in Section 5 and 6, we will give a brief review of related work and conclude the paper.

\section{Background}

Our work is built upon the attention-based neural machine translation model that learns
to align and translate jointly \cite{bahdanauICLR2015:109}, which will be referred to as RNNsearch.

RNNsearch uses a bidirectional RNN \cite{schuster1997bidirectional:110} to encode the source sentence. It consists of two independent RNNs. The forward RNN reads the source sentence from left to right $\overrightarrow{\mathbf{h}}=(\overrightarrow{\mathbf{h}}_1,\dots, \overrightarrow{\mathbf{h}}_{T_x} )$. The backward RNN reads the source sentence from right to left $\overleftarrow{\mathbf{h}}=(\overleftarrow{\mathbf{h}}_1,\dots, \overleftarrow{\mathbf{h}}_{T_x} )$. The representation of the source sentence $\mathbf{h}$ is then defined as the concatenation of $\overrightarrow{\mathbf{h}}$ and $\overleftarrow{\mathbf{h}}$. Each element in $\mathbf{h}$ contains information about the source sentence, focusing on the parts surrounding the corresponding word. 


At decoding time $t$, the attention model uses $\mathbf{s}_{t-1}$ (the RNN states) and $\mathbf{e}_{y_{t-1}}$ (the embedding of previous target word $y_{t-1}$) to ``query'' the encoded $\mathbf{h}$, marks each element of $\mathbf{h}$ a score
$e_{tj} = f(\mathbf{s}_{t-1}, \mathbf{h}_j, \mathbf{e}_{y_{t-1}})$.
The non-linear function $f$ can take on many forms, but we concatenate the three inputs and feed it to a neural network with one hidden layer and tanh as activation function.  
The scores are then normalized to $\{\alpha_{tj}\}$, serving as the weights of  $\{\mathbf{h}_j\{$ to the target word, which then gives the context vector $\mathbf{h}_j$,
$\mathbf{c}_t = \sum_{j=1}^{T_x} \alpha_{tj}\mathbf{h}_j$. The context vector $\mathbf{c}_t$ is then used to update the hidden state of the decoder:
\vspace{-5pt}
\begin{equation}
\label{equ:updatesi}
\mathbf{s}_t = f_u(\mathbf{s}_{t-1}, \mathbf{c}_t, \mathbf{e}_{y_{t-1}}),
\end{equation}
where the function $f_u$ is GRU \cite{cho-EtAlEMNLP2014:114,chung2014empirical:112}.
To predict a target word, the decoder combines $\mathbf{s}_t$, $\mathbf{c}_t$ and $\mathbf{e}_{y_{t-1}}$, feeds to a one-layer MLP with tanh as activation function, followed by a softmax function,
\begin{footnotesize}
\begin{multline}
\label{equ:generatey}
p(y_t=y_i|\mathbf{y}_{<t}, \mathbf{x};\theta) \propto\\ \exp\left\{\mathbf{v}_i^T\mathbf{W}_o\tanh\left(\mathbf{U}_{o}\mathbf{s}_{t-1}+\mathbf{C}_{o}\mathbf{c}_t +\mathbf{V}_{o} \mathbf{e}_{y_{t-1}}\right)\right\},
\end{multline}
\end{footnotesize}
where $\mathbf{W}_o$, $\mathbf{U}_o$, $\mathbf{C}_o$ and $\mathbf{V}_o$ are weight matrices. $\mathbf{v}_i$ is an one-hot indicator vector for $y_i$.

\section{Models}
\label{sec:models}
In this section, we will give more details of phraseNet, more specifically on the preprocessing, encoder and decoders. With two different variants of the mixture models used in decoder, we naturally have two variants of phraseNet, namely phraseNet$_{gate}$ and phraseNet$_{softmax}$.

\subsection{Preprocessing}
\label{sec:preprocessing}
The phrase table $\mathcal{P}$ is a list of rules. Each rule contains a source phrase $\mathbf{p}^{'}_k$ and its translation, a target phrase $\mathbf{p}_k$. Figure \ref{fig: table} gives an example of the phrase table. For simplicity, we first limit ourselves to a subset of rules with the strongest source-target correspondence, which is in contrast to that in phrase-based statistical machine translation (SMT). In SMT, a phrase could have multiple translations with a probability distribution over them. Here we restrict that for each source phrase in $\mathcal{P}$, it has only one translation with probability almost equal to $1$. This will limit the size of $\mathcal{P}$ but can guarantee that the feasible rules are ``reliable'' enough and greatly simplifies the model design and training.

We will introduce the collection of such a phrase table $\mathcal{P}$ later. At the moment, let us assume that we have the table $\mathcal{P}$ which will be utilized to preprocess the sentence pair before encoding. In order to tag a source sentence $\mathbf{x}=(x_1, x_2,\dots, x_{T_x})$, we need to locate its contained phrases. Hence we will find out the rules in $\mathcal{P}$ whose source phrase appears in $\mathbf{x}$, and denote these rules as $\mathcal{P}_{\mathbf{x}}$. To calculate the likelihood during training, we also need to find out the rules in $\mathcal{P}_{\mathbf{x}}$ whose target phrase appears in the target sentence $\mathbf{y}$, denote as $\mathcal{P}_{\mathbf{xy}}$ ($\mathcal{P}_{\mathbf{xy}} \subset \mathcal{P}_{\mathbf{x}}$).

For simplicity, we remove from $\mathcal{P}_{\mathbf{x}}$ the short rules that intersect with the others, which means if the source phrases of two rules are overlapped, we remove the rule whose source phrase has fewer words.  We choose at most $n_{p}$ (a hype-parameter) phrases for each sentence with the maximum coverage. 
\begin{center}
\begin{figure}
\centering
\includegraphics[height=3.0cm]{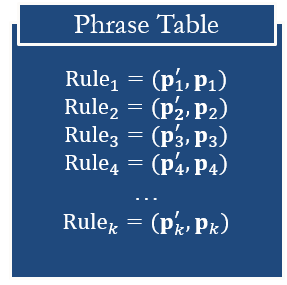}
\caption{The phrase table $\mathcal{P}_{\mathbf{xy}}$. For the $k^{th}$ pair $(\mathbf{p}^{'}_k, \mathbf{p}_k)$, $\mathbf{p}^{'}_k$ stands for the source phrase and $\mathbf{p}_k$ for the target phrase.}
\label{fig: table}
\end{figure}
\vspace{-10pt}
\end{center}

\begin{center}
\begin{figure}
\centering
\includegraphics[width=0.45\textwidth]{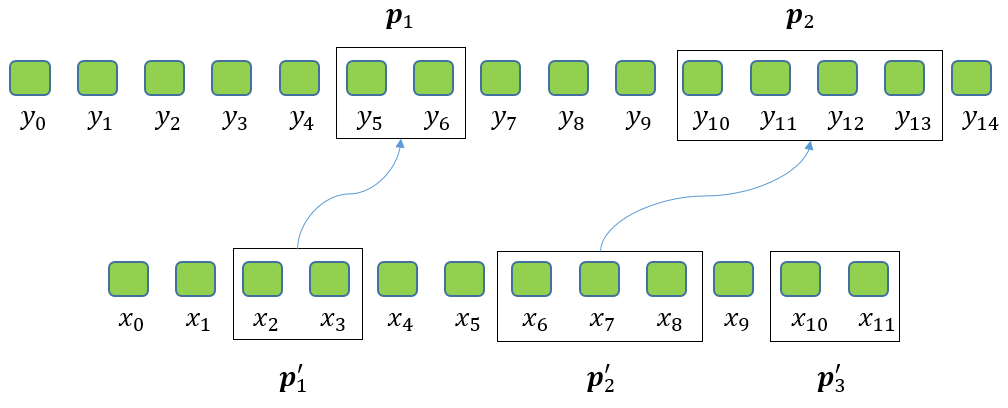}
\caption{Example of a sentence pair being split into groups. $\mathcal{P}_{\mathbf{x}} = \{(\mathbf{p}^{'}_1, \mathbf{p}_1), (\mathbf{p}^{'}_2, \mathbf{p}_2), (\mathbf{p}^{'}_3, \mathbf{p}_3)\}$,  $\mathcal{P}_{\mathbf{xy}} = \{(\mathbf{p}^{'}_1, \mathbf{p}_1), (\mathbf{p}^{'}_2, \mathbf{p}_2)\}$. $\mathcal{W}_x=\{x_0, x_1, x_4, x_5, x_9\}$, $\mathcal{W}_y=\{y_0, y_1, y_2, y_3, y_4, y_7, y_8, y_9, y_{14}\}$.}
\label{fig: sentencepair}
\end{figure}
\end{center}

The non-overlapping rules in $\mathcal{P}_{\mathbf{x}}$ and $\mathcal{P}_{\mathbf{xy}}$ split the words of source and target sentences into groups as showing in Figure \ref{fig: sentencepair}. The words in source sentence $\mathbf{x}$ are split into two groups, phrases $\mathcal{P}_{\mathbf{x}}$ and words not-in-phrases $\mathcal{W}_x$, while the words in target sentence $\mathbf{y}$ are split into two groups, phrases $\mathcal{P}_{\mathbf{xy}}$ and words not-in-phrases $\mathcal{W}_y$.
\vspace{-5pt}
\vspace*{-20pt}
\subsection{Encoder}
\label{sec:encoder}
We use the regular encoder from RNNsearch, but add tags to $\mathbf{h}_i$ \cite{meng-EtAl:113}, $\mathbf{h}^{'}_i= \left[\mathbf{h}_i, ~tag_i\right]$. The tags are used to help the model locate and discriminate different phrases. 

\begin{center}
\begin{figure}[H]
\centering
\includegraphics[width = 0.45\textwidth]{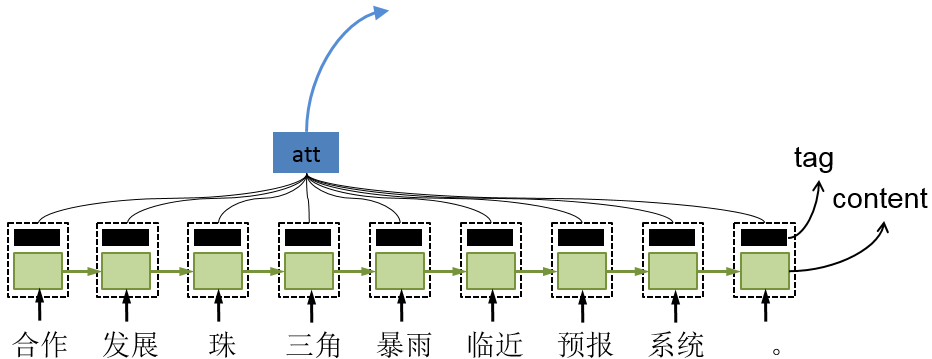}
\caption{The diagram for encoder.}
\label{fig: encoder}
\end{figure}
\end{center}

Each $tag_i$ is an indicator vector with length $n_{p}$. 
For example in Figure \ref{fig: sentencepair}, suppose $n_{p}=5$ and we find three source phrases $\mathbf{p}^{'}_1 = (x_2, x_3), \mathbf{p}^{'}_2 = (x_6, x_7, x_{8}), \mathbf{p}^{'}_3 = (x_{10}, x_{11})$ in sentence $\mathbf{x}$, we concatenate a tag vector $(1,0,0,0,0)$ to each of $\mathbf{h}_2, \mathbf{h}_3$, concatenate a vector $(0,1,0,0,0)$ to each of $\mathbf{h}_6, \mathbf{h}_7, \mathbf{h}_{8}$, concatenate a vector $(0,0, 1,0,0)$ to $\mathbf{h}_{10}$ and $\mathbf{h}_{11}$. For all other not-in-phrase words $\mathcal{W}_x$, we also add a trivial tag vector $(0, 0, 0, 0, 0)$ to their $\mathbf{h}_i$. Figure \ref{fig: encoder} shows an example of such concatenating.
\vspace{-5pt}
\vspace*{-20pt}
\subsection{Decoder}
Unlike the decoder of RNNsearch that only has {\em word mode}, our decoders also have {\em phrase mode}. For a two-word target phrase $\mathbf{p}_t = (y_t, y_{t+1})$, it can either be generated by {\em word mode} (one by one) or {\em phrase mode} (as a whole). In our models, we add upon the RNNsearch another component which has two functions, (1) makes decision between {\em phrase mode} and {\em word mode}, (2) chooses the right target phrase if the decision is {\em phrase mode}. 

With the help of attention model, the new component tries to capture the signals from the encoded representations of a source sentence and translate part of the source sentence (source phrase) directly to the target output as a whole at proper decoding moments. The tags added in Section \ref{sec:encoder} will play an important role in the process. If we view $\mathbf{h}$ as a short-term memory as it changes from sentence to sentence, then the phrase table could be called a phrase memory which is a long-term memory. The decoder queries the short-term memory to choose the segment of source  while it consults the phrase memory to choose the right phrase.

We have two model variants, namely phraseNet$_{gate}$ and phraseNet$_{softmax}$, each corresponding to a different implementation of the mixture model in decoder.
\subsection{phraseNet$_{gate}$}
 \begin{center}
 \begin{figure}[H]
 \centering
 \includegraphics[width=0.48\textwidth]{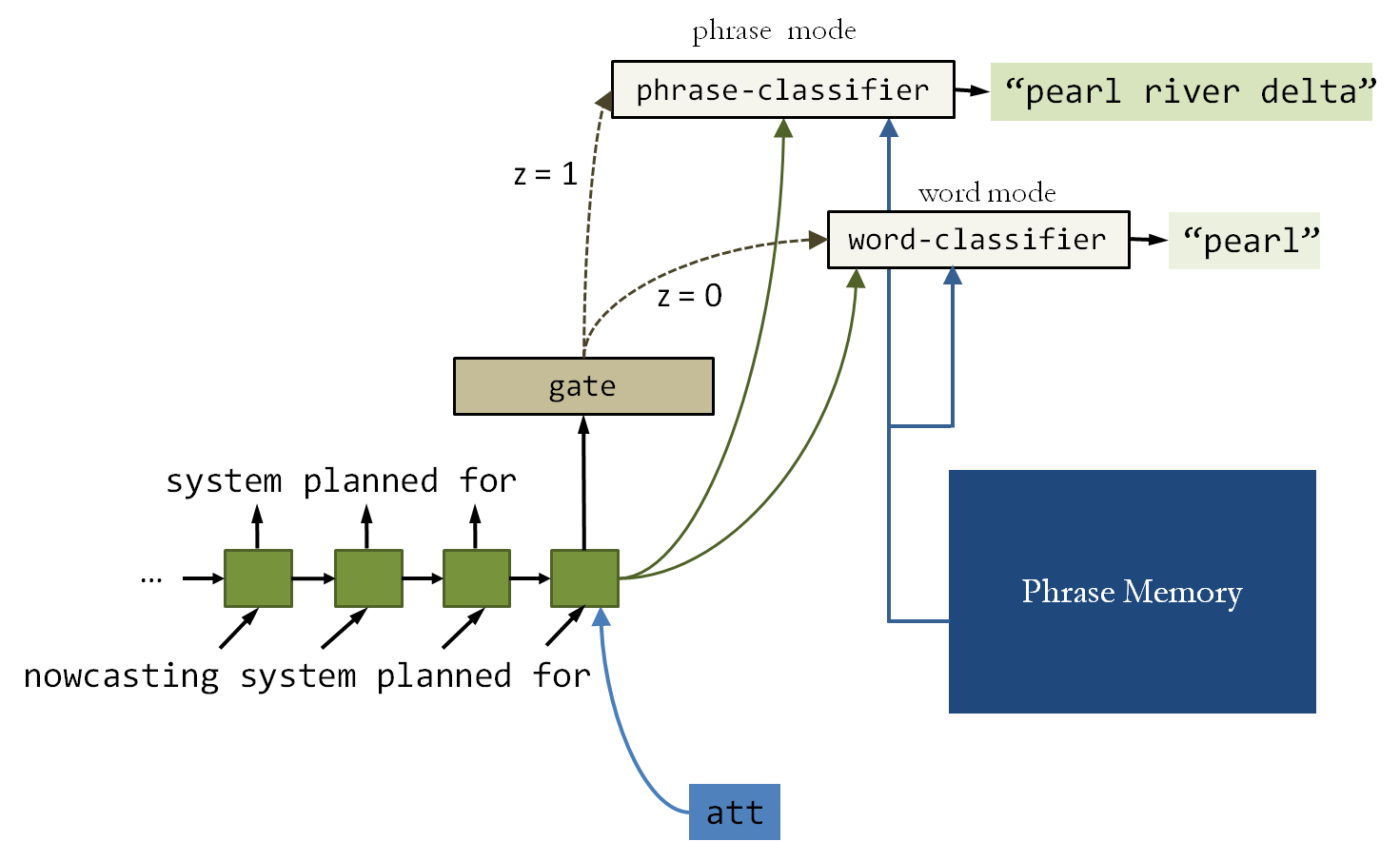}
 \caption{The diagram for phraseNet$_{gate}$.}
 \label{fig: decoder5}
 \end{figure}
 \end{center}
 \vspace*{-20pt}
 
 The decoder for phraseNet$_{gate}$ is illustrated in Figure \ref{fig: decoder5}. In phraseNet$_{gate}$, the decoder will first use a gate $f_z$ to determine the mode at time $t$ by issuing a binary indicator variable ($z_t\in \{0,1\}$), where $0$ represents {\em word mode} and $1$ represents {\em phrase mode}. Then for each mode, it will calculate the word probabilities and phrase probabilities respectively. 
 In word mode, a classifier $f_w$ outputs a  probability distribution over the words in target vocabulary $\mathcal{V}$; while in phrase-mode a phrase classifier $f_{\mathbf{p}}$ determines the probability distribution over the phrases in $\mathcal{P}_{x}$. The final probability of output is given by the probabilities of modes  as well as the probability of individual output generated in each mode. It is a mixture model since a phrase like \texttt{china daily} can be generated in both modes, and the final probability of it is therefore the sum of the probabilities of it being generated from each. For a target phrase that is not in word vocabulary, i.e., it contains  \texttt{UNK}, the probability of that can only be from the phrase mode.
 
 One snapshot of the decoding is the following. Let us suppose that at time $t = 1$, the decoder has generated the word $y_1$ in the {\em word
mode}, and the current state is $\mathbf{s}_1$ , moving to time $t = 2$. With the state $\mathbf{s}_1$ , the attention model (same as in RNNsearch) first generates the context $\mathbf{c}_2$ as a weighted sum of the $\mathbf{h}$ (from encoder). With $\mathbf{s}_1$, $\mathbf{c}_2$ and $\mathbf{e}_{y_1}$, the decoder could move to update the next state $\mathbf{s}_2$ and generating the next word (or phrase). More specifically, in phraseNet$_{gate}$, the state $\mathbf{s}_2$ is updated in the same way as in RNNsearch~(Equation (\ref{equ:updatesi})). The generation of the next word/phrase can be described as follows, let's denote $S_t = (\mathbf{s}_t, \mathbf{c}_t, \mathbf{e}_{y_{t-1}})$:
\begin{enumerate}
\vspace{-5pt}
\item STEP-1: generate the decision variable $z_2$ with 
\begin{eqnarray*}
p(z_2=1|\mathbf{S}_2;\theta) & = & f_z(\mathbf{S}_2)\\
p(z_2=0|\mathbf{S}_2;\theta) & = & 1-f_z(\mathbf{S}_2)
\end{eqnarray*}
\vspace{-20pt}
\item STEP-2a: if $z_2=0$ ({\em word mode}), generate a word based on $\mathbf{s}_2$ with the regular word vocabulary $\mathcal{V}$, same as RNNsearch~(Equation \ref{equ:generatey});
\vspace{-5pt}
\item STEP-2b: if $z_2=1$ ({\em phrase mode}), generate target phrase $\mathbf{p}_j \in \mathcal{P}_{x}$ with probability:
\vspace{-5pt}
\begin{multline*}
p_{\mathbf{p}}(y_2 = \mathbf{p}_j|\mathbf{S}_2, 1; \theta) 
 \propto\\ \exp\left\{\mathbf{u}_j^T\mathbf{W}_p\tanh\left(\mathbf{U}_{p}\mathbf{s}_{2}+\mathbf{C}_{p}\mathbf{c}_2 +\mathbf{V}_{p} \mathbf{e}_{y_{1}}\right)\right\},
\end{multline*}
where $\mathbf{W}_p$, $\mathbf{U}_p$, $\mathbf{C}_p$ and $\mathbf{V}_p$ are weight matrices. $\mathbf{u}_j$ is an one-hot indicator vector for $\mathbf{p}_j$. 
\item STEP-3: calculate the final probabilities and sample the next word (or phrase):
\begin{eqnarray*}
p(y_2=w_i) &= & p(z_2=0|\mathbf{S}_2;\theta)p(w_i|\mathbf{S}_2,0;\theta)\\
p(y_2=\mathbf{p}_j) & = & p(z_2=1|\mathbf{S}_2;\theta)p(\mathbf{p}_j|\mathbf{S}_2,1;\theta)\\
p(y_2) & = & \left[ \begin{array}{c}
p(y_2=w)\\
p(y_2=\mathbf{p})
 \end{array} \right],
\end{eqnarray*}
where the size of $p(y_2)$ is $n_p$ plus the number of words in vocabulary $\mathcal{V}$. The next word or phrase will be sampled according to $p(y_2)$.
If next generation is a phrase, the decoder will go through an ``idle run'' process (Section \ref{sec:idlerun}) to generate words in $\mathbf{p}_2$, after that the decoder replaces the tags $tag_i$ of those source words of $\mathbf{p}^{'}_2$ to all-zero vectors as $\mathbf{p}^{'}_2$ has already been decoded.  
\end{enumerate}

Similar to \cite{gulcehre2016pointing:107}, in phraseNet$_{gate}$, $f_z$ is a three-layered neural network using noisy-tanh activation for the first two layers \cite{gulcehre2016noisy:105} and we add residual connection \cite{he2015deep:106} from the first layer to the second hidden layer, The output layer uses sigmoid as activation function. 

\subsection{phraseNet$_{softmax}$}
\begin{center}
 \begin{figure}[h!]
 \centering
 \includegraphics[width=0.5\textwidth]{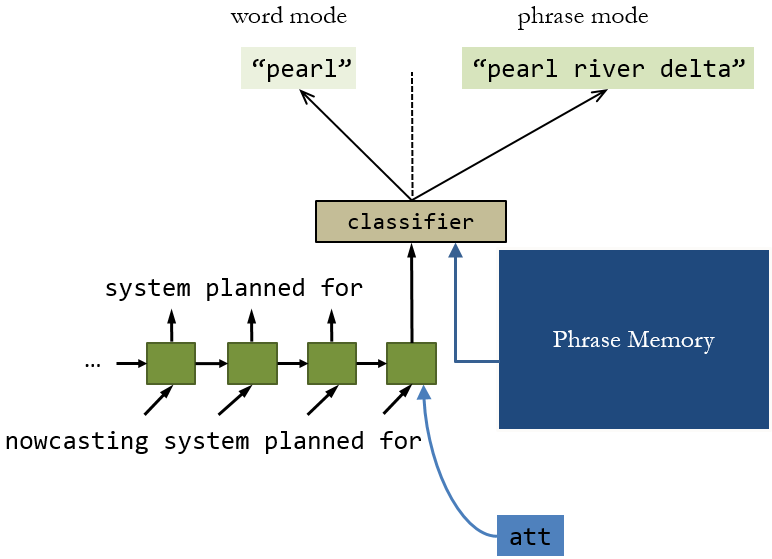}
  \caption{The diagram for phraseNet$_{softmax}$.}
 \label{fig: decoder6}
 \end{figure}
 \end{center}

With phraseNet$_{gate}$ the decision of phrase mode is made before seeing the actual content of the target phrase, which fails to make use of the language model and semantic relevance on the target side. To address this drawback we devise phraseNet$_{softmax}$, which takes the candidate phrases and candidate words in the same softmax, as illustrated in Figure~\ref{fig: decoder6}. To do this, all the phrases need to  embedded as vectors, where the embedding model is also learned in the NMT training. It is also worth to mention that phraseNet$_{softmax}$ can  potentially handle the case where one source phrase may correspond to multiple candidate target phrases, since the decoder can distinguish them based on their content. This modeling advantage, however, will not be explored in this paper.

 \vspace*{-20pt}
Given a rule $(\mathbf{p}^{'}_k, \mathbf{p}_k)$, we have several choices to calculate its embedding. 
In this paper, we choose to use a separate backward RNN to encode $\mathbf{p}_k$ and choose the last state as the embedding for it. That way, the embedding will keep more information of the first word of 
$\mathbf{p}_k$, therefore facilitate a potential language model in scoring ($y_{i-1}, \mathbf{s}_i, \mathbf{p}_k$). 



Suppose that at time $t = 1$, the decoder has generated the word $y_1$ in the {\em word
mode}, and the current state is $\mathbf{s}_1$ , moving to time $t = 2$. 
With the state $\mathbf{s}_1$ , the decoder makes an attentive read to  $\mathbf{h}$ to obtain $\mathbf{c}_2$. The state $\mathbf{s}_2$ is updated in the same way as in Equation (\ref{equ:updatesi}). The generation of the next word/phrase can be described as follows:
\begin{enumerate}
\vspace{-5pt}
\item STEP-1: calculate the word score for each word in $\mathcal{V}$,
\begin{eqnarray*}
\psi_{w_i} = \mathbf{v}_i^T \mathbf{W}_{w}\tanh(\mathbf{U}_{w}\mathbf{s}_{2}+\mathbf{C}_{w}\mathbf{c}_2 +\mathbf{V}_{w} \mathbf{e}_{y_{1}}),
\end{eqnarray*}
where $\mathbf{W}_{w}$, $\mathbf{U}_{w}$, $\mathbf{C}_{w}$ and $\mathbf{V}_{w}$ are weight matrices.
\vspace{-5pt}
\item STEP-2: calculate the phrase score for each phrase in $\mathcal{P}_{x}$
\begin{equation*}\footnotesize
\psi_{\mathbf{p}_j} =
\mathbf{W}_{q}\tanh(\mathbf{U}_{q}\mathbf{s}_{2}+\mathbf{C}_{q}\mathbf{c}_2 +\mathbf{V}_{q} \mathbf{e}_{y_{1}}+\mathbf{R}_{q}\mathbf{e}_{\mathbf{p}_j}),
\end{equation*}
where $\mathbf{e}_{\mathbf{p}_j}$ is the embeddings of rule $(\mathbf{p}^{'}_j, \mathbf{p}_j)$. $\mathbf{W}_{q}$, $\mathbf{U}_{q}$, $\mathbf{C}_{q}$, $\mathbf{V}_{q}$ and $\mathbf{R}_{q}$ are weight matrices.
\vspace{-5pt}
\item STEP-3: calculate the probabilities of all words and phrases through softmax
\begin{eqnarray*}
p(y_2|\mathbf{s}_2, \mathbf{c}_2, \mathbf{e}_{y_1}) &=& \text{softmax}(\left[\mathbf{\psi}_w, \mathbf{\psi}_{\mathbf{p}}\right])
\end{eqnarray*}
In the softmax, the phrases will compete directly with words, which is different from the phraseNet$_{gate}$ where phrase probabilities and word probabilities are calculated independently. 
While phraseNet$_{gate}$ has difficulties in calculating the scores for each phrase, phraseNet$_{softmax}$ has the flexibility to adapt to the new setting with embeddings. If the choice is a phrase, the decoder will go through ``idle run'' process and the tags of those source words of the chosen phrase will be set to all-zero.
\end{enumerate}

 
\subsection{Idle Run for Multi-word Phrases}
\label{sec:idlerun}
Our decoder is vastly different from that of RNNsearch, but it is still generally built on the basic word-by-word decoding mechanism. To further accommodate the phrase mode in which multiple words are generated all at once, we introduce the ``idle run''. Basically, if at time $t$ a multiple-word phrase is chosen, the decoding RNN will run exactly the same way as in word mode with regard to state update and attention, only that the generation of words in the rest of phrase is pre-determined at $t$. 

This process is called \emph{idle run}, which can be illustrated through the following example. In Figure \ref{fig: sentencepair}, if at time $t=5$, the decoder decides to go with {\em phrase mode} and generate $\mathbf{p}_1$, the decoder will not really output $\mathbf{p}_1=(y_5, y_6)$ at once. To keep the updating of $\mathbf{s}_t$ (Equation (\ref{equ:updatesi})), the decoder will first output $y_5$ and use $\mathbf{e}_{y_5}$ and other required elements to update $\mathbf{s}_5$ to $\mathbf{s}_{6}$, uses $\mathbf{s}_{6}$ to generates $y_{6}$. With $\mathbf{e}_{y_6}$ and other required elements, the decoder will update the state to $\mathbf{s}_{7}$ and at the time $t=7$, the decoder starts to make its next decision, {\em phrase mode} or {\em word mode} for the coming words. During the output of $\mathbf{p}_1$, the decoder does not need to make decisions or sample words as it is already in one {\em phrase mode}.

\subsection{The Probabilistic Model for Phrases}
Given a target phrase $\mathbf{p}=\{y_t, y_{t+1}, y_{t+2}\}$, in principle, its words could be chosen either from vocabulary $\mathcal{V}$ or entirely retrieved from phrase table $\mathcal{P}$. So in general each word is potentially generated from a mixture probability model. In the case that there are out-of-vocabulary words (\texttt{UNKs}) in the phrases, which are faily common in practice, the mixture model degenerates to phrase mode only. 

For an unified notation, we introduce an indicator variable $I_{unk}$ into the mixture probability model, which is summarized as follows, 
\begin{eqnarray}
\begin{aligned}
 p(y_t, y_{t+1}, & y_{t+2}|\mathbf{y}_{<t}, \mathbf{x}; \theta)\\
=& I_\texttt{unk}\times\prod_{i=t}^{t+2}p(z_i=0, y_i| S_i;\theta)\\
+& p(z_t=1, \mathbf{p}_t=\mathbf{p} |S_t;\theta).
\end{aligned}
\label{formu:mixture}
\end{eqnarray}
where $I_{unk}=1$ means there is no \texttt{UNK}s in the phrase, and $0$ otherwise.

For phraseNet$_{gate}$, $p(z_t=0, y_t| S_t;\theta)$ factorizes into $p(z_t=0|S_t;\theta)p(y_t|S_t, 0; \theta)$, and $p(z_t=1, \mathbf{p}_t=\mathbf{p} |S_t;\theta)$ factorizes into $p(z_t=1|S_t;\theta)p_{\mathbf{p}}(\mathbf{p}_t=\mathbf{p}|S_t, 1; \theta)$.
For phraseNet$_{softmax}$, there is no explicit variable $z_t$, the indicator of mode is implicitly absorbed into the choice of words and phrases. The probability $p(z_t=0, y_t| S_t;\theta)$ can therefore be re-written as $p(y_t| S_t;\theta)$ and $p(z_t=1, \mathbf{p}_t=\mathbf{p} |S_t;\theta)$ as $p(\mathbf{p}_t=\mathbf{p} |S_t;\theta)$. 

For normal words that are not part of phrases $\mathcal{W}_y$, they can only be generated by {\em word mode}, which is the same as RNNsearch.

Given a pair of source and target sentence $\mathbf{x}=(x_1, x_2,\dots, x_{T_x})$ and $\mathbf{y}=(y_1, y_2,\dots, y_{T_y})$, the probability of this pair of sentences is:
\begin{small}
\begin{equation*}
p(\mathbf{y}|\mathbf{x}; \theta) = \prod_{y_i \in \mathcal{W}_y} p(y_i|\mathbf{y}_{<i}, \mathbf{x}; \theta) \prod_{p_j \in \mathcal{P}_{xy}} p(\mathbf{p}_j|\mathbf{y}_{<j}, \mathbf{x}; \theta),
\end{equation*}
\end{small}
here $p(\mathbf{p}_j|y_{<j}, \mathbf{x}; \theta)$ refers to the mixture probability (Equation (\ref{formu:mixture})) of output the words in $\mathbf{p}_j$. For a given batch of the source and
target sequences $\{X\}_N$ and \{$Y_N \}$, the objective is to minimize the negative log-likelihood:
\begin{eqnarray*}
\mathcal{L} = -\frac{1}{N}\sum_{k=1}^{N}p(\mathbf{y}^{(k)}|\mathbf{x}^{(k)}; \theta)
\end{eqnarray*}

\section{Experiments}
\label{sec:length}
We report our empirical study on applying phraseNet$_{gate}$ and phraseNet$_{softmax}$  to Chinese-to-English translation, and comparing it against RNNsearch and SMT models.
\subsection{Phrase table $\mathcal{P}$}
\begin{center}
\begin{figure}[h!]
 \centering
 \includegraphics[height=3.3cm]{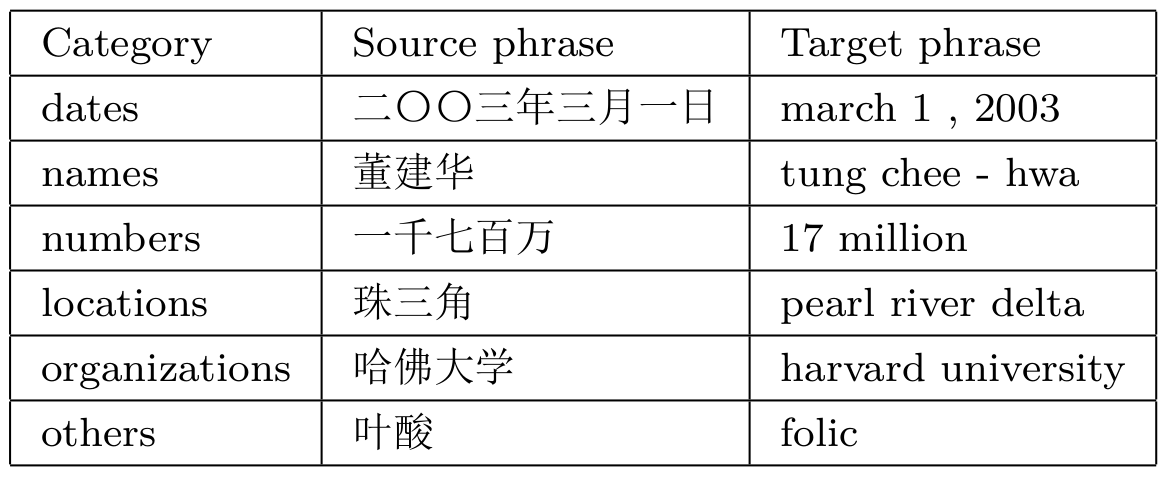}
  \caption{Examples of phrase for each category.}
 \label{fig:phrasetableexample}
 \end{figure}
\end{center}

As mentioned before, when we design our model, our definition for ``phrase'' is different from that used in phrase-based statistical machine translation. For each source phrase, our models only support an unique translation (target phrase). Therefore, we only choose those phrase pairs that the source phrase almost always translates to the target phrase. We also hope the contexts for the source phrases are relatively fixed so that the models can learn the patterns of translation easier. With these considerations, we focus our attention on five categories of phrases: dates, names, numbers, locations and organizations. Apart from these five categories, we also collect some other phrases that fulfil our requirements. Figure \ref{fig:phrasetableexample} shows several examples of our Phrase table.
\vspace*{-20pt}
The phrase pairs are collected from several sources. The first source consists of extracted phrase pairs from a bi-lingual corpus using the method described in \cite{meng2014modeling:100}, \cite{ren2009improving:101} and \cite{frantzi2000automatic:102}. The second source is the LDC dictionary. The third source is from proper nouns dictionaries, which contain many commonly used Chinese-to-English translation pairs for proper nouns. We have also generated some Chinese-to-English phrase translations, especially dates, numbers and Chinese names, by predefined rules. There are two formats of Chinese names, Mandarin names and Cantonese names, both of their English counterparts could be generated according to their pronunciation rules. Numbers can also be generated by predefined rules, like ``1345 $\rightarrow$ 1,345''. Using these rules, we transform several formats of Chinese numbers to English numbers.

\subsection{Setup}
Our training data contains 1.25M sentence
pairs obtained from LDC corpora\footnote{The corpora include LDC2002E18, LDC2003E07, LDC2003E14, Hansards portion of LDC2004T07, LDC2004T08 and LDC2005T06.}, with 27.9M Chinese words and 34.5M English words respectively. We use NIST 2002 (NIST02) dataset as our development set, and the NIST 2003 (NIST03), NIST 2004 (NIST04), NIST 2005 (NIST05), 2006 (NIST06) and 2008 (NIST08) datasets as our test
sets. The case-insensitive 4-gram NIST BLEU score (Papineni et al.2002) is used
as our evaluation metric. 

In training the neural networks, we limit the source
and target vocabularies to the most frequent 16K words (one of the
words is reserved for the unknown words (UNK)) in Chinese and English,
covering approximately $95.8\%$ and $98.3\%$ of the two corpora respectively. We train each model
with the sentences of length up to 50 words in training data. The word embedding
dimension is 620 and the size of a hidden layer is 1000. We set $n_{p}$ as 10.
In both the RNNsearch and our models, we adopt the coverage models introduced in \cite{tu2016modeling:103} to mitigate the problem of over-translation.

We compare our models with state-of-the-art SMT and RNNsearch:
\begin{enumerate}
\vspace{-5pt}
\item Moses (Koehn et al.2007): an open source phrase-based translation system
with default configuration and a 4-gram language model trained on the target
portion of training data;
\vspace{-5pt}
\item RNNsearch (Bahdanau et al.2015): an attentional NMT model with default setting.\footnote{We use the code from (\url{https://github.com/kyunghyuncho/dl4mt-material}) with minor modifications.}

\end{enumerate}

\subsection{Translation Performance}
Table \ref{tab:bleuscore} shows the translation performances measured in BLEU score. Clearly both the proposed phraseNet$_{gate}$ and phraseNet$_{softmax}$ significantly improves the translation quality in all cases. More specifically, On average, phraseNet$_{gate}$ yields about $3.45$ BLEU score improvement over our baseline, phraseNet$_{softmax}$ yields about $2.13$ BLEU score improvement over our baseline. Also, RNNsearch with expanded  vocabulary (30K words) is 1.65 BLEU behind phraseNet$_{gate}$. Surprisingly, phraseNet$_{softmax}$ comes behind phraseNet$_{gate}$, despite its potential ability to take the content of the target phrase into the decision. We conjecture that this might be due to the difficulty in directly comparing the scores from two different types of scoring functions  in the same pool of softmax. 

\begin{table*}[bt]
\centering
\footnotesize
\begin{tabular}{|l||l||l|l|l|l|l|l|}
\hline
{\bf Models} & {\bf NIST02}& {\bf NIST03}& {\bf NIST04}& {\bf NIST05}& {\bf NIST06}& {\bf NIST08} & {\bf Ave.} \\\hline
Moses        & 33.41   & 31.61   & 33.48   & 30.75   & 31.07   & 23.37 & 30.06\\\hline
RNNSearch (16K)    & 34.96   & 32.19   & 33.85   & 30.79   & 30.32   & 22.13 & 29.86\\\hline
RNNSearch (30K)    & 36.04   & 33.96   & 35.82   & 33.05   & 31.88   & 23.61 & 31.66\\\hline
phraseNet$_{gate}$ (16K)& \bf{37.68}   & \bf{36.01}   & \bf{37.69}   & \bf{34.61}   & \bf{32.70}   & \bf{25.52} & \bf{33.31}\\\hline
phraseNet$_{softmax}$ (16K) & 36.60   & 34.07   & 35.93   & 33.37   & 31.96   & 24.62 & 31.99\\\hline
\end{tabular} 
\caption{Evaluation of translation quality, where we use boldface digits to denote the best performance. }
\label{tab:bleuscore}
\end{table*}

It is also reasonable to doubt that our models are just generate the target phrases without considering the positions, as this will also (almost surely) increase the 1-gram and 2-gram BLEU scores and hence increase the final BLEU scores.  To further verfy this, Table \ref{tab:bleuscore4} compares our models with RNNsearch measured in 4-gram BLEU score, which capture overlapping of generated targets and reference on longer segments.

Our models, especially phraseNet$_{gate}$, still perform better than RNNsearch, incidating that the phrases are put into the right places.
\begin{table*}
\centering
\footnotesize
\begin{tabular}{|l||l||l|l|l|l|l|l|}
\hline
{\bf Models} & {\bf NIST02}& {\bf NIST03}& {\bf NIST04}& {\bf NIST05}& {\bf NIST06}& {\bf NIST08} & {\bf Ave.} \\\hline
RNNSearch (16K)   & 16.89   & 15.64   & 17.77   & 16.02   & 15.26   & 10.16 & 14.97\\\hline
phraseNet$_{gate}$ (16K) & 18.97   & 17.95   & 19.11   & 17.21   & 16.14   & 11.92 & 16.47\\\hline
phraseNet$_{softmax}$ (16K) & 17.72   & 16.23 & 17.94 & 16.42 & 15.53 & 10.98  & 15.42\\\hline
\end{tabular} 
\caption{Evaluation of translation quality in 4-gram BLEU score.}
\label{tab:bleuscore4}
\end{table*}
\begin{center}
\begin{figure*}[t!]
 \centering
 \includegraphics[width=0.85 \textwidth]{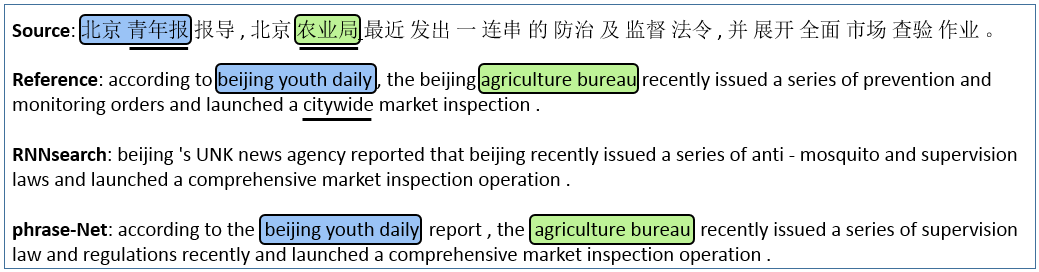}
 \includegraphics[width=0.85 \textwidth]{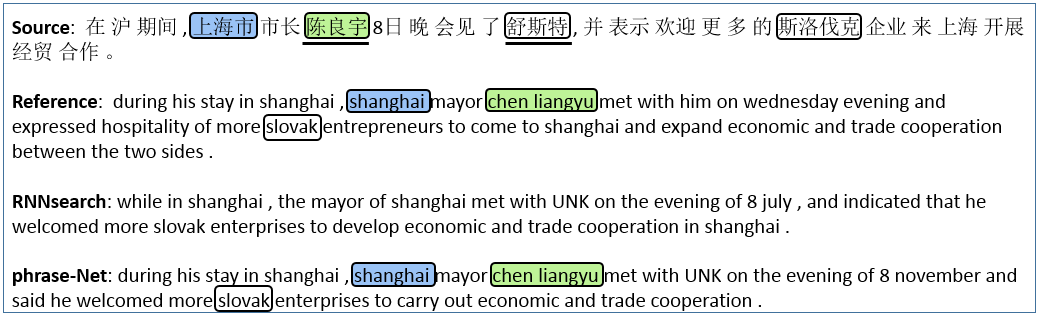}
  \caption{Example of phraseNet$_{gate}$ on test sets compared with RNNsearch. Word segmentation is applied on the input, where underlined are UNK words. The phrases highlighted by boxes (with or without colors) are those phrases in our phrase table. The highlighted phrases without colors are phrases generated by {\em word mode} or not generated.}
 \label{fig:translateexample}
 \end{figure*}
\end{center}
\vspace*{-20pt}
\subsection{Samples of Translation}
We also give two examples from test set comparing our phraseNet$_{gate}$ with RNNsearch, and more examples can be found in supplementary materials.  As demonstrated through those examples, when there are phrases found in the source sentences, phraseNet$_{gate}$ has a better chance to generate the corresponding target phrases correctly at proper locations. This could happen when the source phrases consist words all in the vocabulary, but more frequently when there are UNK words there, showing that phraseNet$_{gate}$ is also a strong model to solve the UNK problem. Another interesting observation, e.g., the second example in Figure~\ref{fig:translateexample}  is that for some common phrases  phraseNet sometimes ignores the suggestion of phrase mode, but still generate the entire phrase correctly from its word mode. This shows phraseNet maintains a healthy and flexible balance between word and phrase mode.
\vspace{-7pt}
\section{Related Work}
\label{sec:rework}
\vspace{-7pt}
Probably the work that is closest to phraseNet is the recently proposed Neural Generative QA (genQA) \cite{yin2015neural:115}, where a set of triples are stored in a QA memory, and a neural network queries this memory for words to use in generating the answer. More specifically, phraseNet$_{gate}$ has the same gating strategy as in genQA. Still, phraseNet is different from that in several important ways: 1) phraseNet can handle multiple phrases in one sentence, and 2) phraseNet can generate multi-word expression.

The softmax with multiple modes in phraseNet$_{softmax}$ is very similar to the recently proposed CopyNet~\cite{JiataoGu:116}. But the generative mode in CopyNet still follows a strict word-by-word fashion and therefore a soft-decision between modes has to be made for each mode.  In a similar way, phraseNet is related to  \cite{gulcehre2016pointing:107} and \cite{cheng2016neural:119}.
\vspace{-7pt}
\section{Conclusions and Future Work}
\vspace{-7pt}
We propose a neural machine translator which can leverage an external phrase memory, and empirically show its efficacy on Chinese-English translation.

\bibliography{pn_nmt_emnlp2016}
\bibliographystyle{pn_nmt_emnlp2016}

%
%

\end{document}